\pdfoutput=1

\documentclass[11pt]{article}

\usepackage{EMNLP2023}
 
\usepackage{bbm}
\usepackage[]{mdframed}
\usepackage{xcolor}
\usepackage{adjustbox}
\usepackage{comment}
\usepackage{multirow}

\usepackage{physics}
\usepackage{array}
\usepackage{color,soul}
\usepackage{placeins}
\usepackage{amssymb}
\usepackage{booktabs, multirow, array, makecell, caption}

\usepackage{xcolor}
\usepackage{times}
\definecolor{aqua}{rgb}{0.0, 1.0, 1.0}
\usepackage{latexsym}

\usepackage{tabularx}
\usepackage{hyperref}
\usepackage{pifont}
\usepackage{verbatim}
\usepackage{soul}
\newcommand{\hlc}[2][yellow]{{%
    \colorlet{foo}{#1}%
    \sethlcolor{foo}\hl{#2}}%
}

\usepackage{subcaption}
\usepackage{mwe}

\usepackage{mathtools}
\usepackage{lipsum}
\usepackage{float}
\newcolumntype{b}{X}
\newcolumntype{s}{>{\hsize=.5\hsize}X}
\usepackage{times}
\usepackage{latexsym}
\usepackage[T1]{fontenc}
\usepackage[utf8]{inputenc}
\usepackage{microtype}
\definecolor{aqua}{rgb}{0.0, 1.0, 1.0}
\usepackage{inconsolata}
\usepackage{CJKutf8}
\usepackage{graphicx}
\usepackage{array}
\usepackage{arydshln}
\usepackage{multirow}

\usepackage{bm}
\usepackage{amssymb}
\usepackage{amsmath}
\usepackage{makecell}
\usepackage{wrapfig}
\usepackage{lipsum}  
\usepackage{etoolbox}
\usepackage{comment}
\usepackage{enumitem}

\usepackage{booktabs}
  
\usepackage{microtype}
\usepackage{csquotes}
\usepackage{bm}
\usepackage{amsfonts,mathtools}
\usepackage{soul}
\usepackage{adjustbox}
\usepackage[capitalize]{cleveref}
\usepackage{xcolor}
\usepackage{hyperref}
\usepackage{appendix}
\usepackage{svg}
\AtBeginEnvironment{appendices}{\crefalias{section}{appendix}}
\BeforeBeginEnvironment{appendices}{\clearpage}

\Crefname{appsec}{appendix}{appendices}

\usepackage{caption}
\usepackage{subcaption}
\usepackage{enumerate}
\usepackage{scalerel}
\usepackage[ruled,vlined]{algorithm2e}
\usepackage{linguex}
\usepackage{relsize}
\usepackage{array}
\usepackage{booktabs}
\usepackage{array}
\usepackage{makecell}
\usepackage{relsize}
\usepackage{multirow}

\usepackage[font=small,labelfont=bf]{caption}

\usepackage{cancel}
\usepackage{inconsolata}

\usepackage[utf8]{inputenc}
\definecolor{darkorange}{rgb}{1, 0.549, 0}
\BeforeBeginEnvironment{appendices}{\clearpage}

\usepackage{todonotes}

\title{EPA: Easy Prompt Augmentation on Large Language Models \\ via Multiple Sources and Multiple Targets}

\author{Hongyuan Lu, Wai Lam\\
  The Chinese University of Hong Kong \\
  \texttt{\{hylu,wlam\}@se.cuhk.edu.hk} \\}

\begin{document}
\maketitle
\begin{abstract}
Large language models (LLMs) have shown promising performance on various NLP tasks via task prompting. And their performance can be further improved by appending task demonstrations to the head of the prompt. And usually, a better performance can be achieved with more demonstrations. However, asking the users to write the demonstrations can be cumbersome. As a simple yet cost-effective workaround, this paper proposes a novel method called EPA (\textbf{E}asy \textbf{P}rompt \textbf{A}ugmentation)\footnote{While this paper considers augmenting prompts via demonstrations, we name it EPA as the name EDA is already taken by a well-known NLP method \citep{wei-zou-2019-eda}.} that effectively minimizes user efforts in writing demonstrations while improving the model performance at the same time. EPA achieves these goals by automatically augmenting the demonstrations with multiple sources/targets, where each of them paraphrases each other. This is well motivated as augmenting data via paraphrasing effectively improves neural language models. EPA thus employs paraphrasing as an augmentation method for in-context learning. Extensive experiments indicate that EPA effectively improves both NLU and NLG tasks, covering from natural language inference to machine translation in translating tens of languages.\footnote{Code and data will be released upon publication.}

\end{abstract}
\section{Introduction}
Large language models (LLMs) possess the ability to carry out various understanding and generation tasks from natural language inference to machine translation \citep{NEURIPS2020_1457c0d6,2021arXiv211210668L,2022arXiv220501068Z,2022arXiv220111903W,2023arXiv230304048W}. Such an ability is closely related to in-context learning \citep{rubin-etal-2022-learning, zhang-etal-2022-active}. In-context learning prepends one or more source/target pairs (namely demonstrations) to the head of the requests. This effectively improves the downstream task performance. However, most of the scientific research constraints to the situation where those demonstrations are always available \citep{2022arXiv220212837M,2023arXiv230100234D}. 
\par
Yet, the above-mentioned situation is unrealistic, as human annotation is expensive, and there is no guarantee that there always exist enough demonstrations. This is an important consideration, especially for commercial products, where we would like to reduce user efforts in writing demonstrations. This subsequently leads to a better user experience.
\par 
In contrast, this paper considers a realistic situation where only a few demonstrations are available. We propose a simple yet effective framework called EPA (\textbf{E}asy \textbf{P}rompt \textbf{A}ugmentation) that creates multiple sources and multiple targets respectively where they paraphrase each other.
\par
EPA is also well-motivated by the prior works \citep{gao-etal-2020-paraphrase,lu-lam-2023-pcc} that doing data augmentation via paraphrasing improves the neural language model during the model training stage. EPA considers paraphrasing both sources and targets to enhance in-context learning.
In summary, this paper makes three key contributions:
\begin{itemize}
\setlength\itemsep{0em}
    \item We propose a novel method that is easy to use called EPA, aiming at improving the performance of LLMs via in-context learning.
    \item EPA yields promising performance on various NLP tasks, covering from natural language inference to machine translation.
    \item In-depth analyses indicate that EPA brings good improvement, and naively copying the demonstration degrades the performance.
\end{itemize}
EPA is a simple yet effective method for improving LLMs. Therefore, we hope that EPA can benefit our community not only in terms of research but also in developing commercial products.

\section{Easy Prompt Augmentation}
\paragraph{Motivation} EPA is inspired by the fact that paraphrasing can improve the evaluation of natural language generation tasks by enriching the target side of the test instances \citep{thompson-post-2020-automatic,bawden-etal-2020-study,2023arXiv230515067T}. This is due to the fact that one meaning can usually be represented in several sentences in terms of natural language, and multiple targets can provide a precise evaluation of the actual meaning of the generations.
\par 
Meanwhile, the rise of LLMs provides a unified solution to different language-related tasks via the interface of prompts. Prompting can be further enhanced by in-context learning, where source/target pairs can be prepended to the head of the prompts to serve as demonstrations. Yet, the construction of demonstrations usually requires human efforts, which can be clumsy. This is especially inconvenient for commercial products, where we would like to reduce user efforts as much as possible.
\paragraph{Our Approach} Motivated by the need to automatically construct demonstrations instead of human efforts, this paper proposes and investigates the effectiveness of paraphrasing when a small set of demonstrations are already available. Specifically, EPA assumes there are a small set of demonstrations that are already. EPA then considers paraphrasing the demonstrations on both \textbf{source side} and \textbf{target side} to create new demonstrations for in-context learning. Formally, assuming we have one demonstration, the traditional in-context learning feeds the following text concatenation into LLMs:
\[
[x_d,y_d,x],
\]
where $x_d$ and $y_d$ represent the source and target sentence of the demonstration pair. $x$ represents the actual prompt that we hope LLMs solve. EPA first paraphrases $x_d$ and $y_d$ into additional demonstrations $x_{d1}$, $x_{d2}$, $x_{d3}$,..., $x_{dn}$ paired with $y_{d1}$, $y_{d2}$, $y_{d3}$,..., $y_{dn}$ where $x$s and $y$s represents the same meaning. $n$ represents the number of additional paraphrases. EPA then feeds the following text concatenation into LLMs:
\[
[x_d,y_d,x_{d2},y_{d2},x_{d3},y_{d3},...,,x_{dn},y_{dn},x]
\]
The whole EPA framework aims at improving the performance of LLMs while minimizing human efforts in writing demonstrations simultaneously. Therefore, EPA can reduce user efforts and improve the user experience for commercial products.
\section{Experimental Setup}
\subsection{Datasets}
We conduct extensive experiments to verify the effectiveness of EPA on a variety of both understanding and generation tasks:
\begin{itemize}
\setlength\itemsep{0em}
    \item \textbf{Machine Translation} We randomly select 45 high-resourced and low-resourced languages from FLORES-200 \citep{nllb2022}, which is an MT dataset that contains about 200 languages with 1,012 parallel sentences included in the dataset, which were extracted from English Wikipedia covering a variety of topics and domains. The sentences were curated manually by professional translators from English into other languages. We report on translating all the languages in FLORES-200 from English, with the complete dev-test that has about 1,000 instances.
    \item \textbf{Dialogue Summarization} SAMSum \citep{gliwa-etal-2019-samsum} is a dialogue summarization dataset created manually created by linguists who are fluent in English. We use the test set that contains 819 instances for evaluations.
    \item \textbf{Paraphrasing} Quora
Question Pairs (QQP) is a paraphrasing dataset that requires generating an alternative surface form in the same language that maintains the original semantic meaning. We use the preprocessed version from \citet{2022arXiv221008933G} that contains about 4.65k test instances.
    \item \textbf{Natural Language Inference} The SNLI corpus \citep{bowman-etal-2015-large} is a collection of manually-written English sentence pairs labelled for balanced classification between entailment, contradiction, and neutral relationship. We use the test set that contains about 10k instances. The MNLI corpus \citep{N18-1101} is a collection of manually-written sentence pairs annotated with textual entailment information. It is modelled based on the SNLI corpus. MNLI covers a range of categories of spoken and written text. MNLI also supports a distinctive evaluation of cross-genre generalization. We use the matched validation set with about 9.82k instances.
\end{itemize}
\begin{table*}[thb!]
\tiny
\centering
    \setlength\tabcolsep{6pt}
\setlength\aboverulesep{0pt}\setlength\belowrulesep{0pt}
\setcellgapes{0pt}\makegapedcells
\begin{tabular}{l|cc|l|cc|l|cc|l|cc|l|cc}
\hline
\noalign{\vskip 1mm}  
\textbf{Language} & \textbf{GPT} & \textbf{EPA} & \textbf{Language} & \textbf{GPT} & \textbf{EPA} & \textbf{Language} & \textbf{GPT} & \textbf{EPA} & \textbf{Language} & \textbf{GPT} & \textbf{EPA} & \textbf{Language} & \textbf{GPT} & \textbf{EPA}\\
\noalign{\vskip 1mm}  
\hline
\noalign{\vskip 1mm}  
acm\_Arab&23.97&\hlc[aqua]{\textbf{26.28}}&afr\_Latn&51.97&\hlc[aqua]{\textbf{70.16}}&als\_Latn&59.41&\hlc[aqua]{\textbf{63.04}}&arz\_Arab&24.84&\hlc[aqua]{\textbf{29.38}}&asm\_Beng&12.87&\hlc[aqua]{\textbf{27.03}}\\azb\_Arab&1.81&\hlc[aqua]{\textbf{2.84}}&bak\_Cyrl&18.99&\hlc[aqua]{\textbf{39.03}}&bjn\_Arab&2.28&\hlc[aqua]{\textbf{2.38}}&crh\_Latn&26.38&\hlc[aqua]{\textbf{27.69}}&dan\_Latn&57.85&\hlc[aqua]{\textbf{64.02}}\\eus\_Latn&45.86&\hlc[aqua]{\textbf{47.10}}&ewe\_Latn&28.11&\hlc[aqua]{\textbf{40.42}}&fin\_Latn&67.78&\hlc[aqua]{\textbf{75.32}}&fuv\_Latn&12.24&\hlc[aqua]{\textbf{18.16}}&gaz\_Latn&30.76&\hlc[aqua]{\textbf{36.62}}\\guj\_Gujr&23.12&\hlc[aqua]{\textbf{41.61}}&jpn\_Jpan&30.00&\hlc[aqua]{\textbf{37.57}}&kas\_Deva&1.58&\hlc[aqua]{\textbf{9.09}}&khk\_Cyrl&37.40&\hlc[aqua]{\textbf{65.40}}&khm\_Khmr&24.17&\hlc[aqua]{\textbf{28.47}}\\kik\_Latn&20.44&\hlc[aqua]{\textbf{35.82}}&kir\_Cyrl&21.24&\hlc[aqua]{\textbf{36.73}}&kmb\_Latn&8.78&\hlc[aqua]{\textbf{21.23}}&kor\_Hang&20.54&\hlc[aqua]{\textbf{27.69}}&lij\_Latn&45.80&\hlc[aqua]{\textbf{48.41}}\\lim\_Latn&27.70&\hlc[aqua]{\textbf{28.23}}&lmo\_Latn&16.79&\hlc[aqua]{\textbf{17.86}}&ltg\_Latn&22.93&\hlc[aqua]{\textbf{27.63}}&lug\_Latn&24.31&\hlc[aqua]{\textbf{29.76}}&mai\_Deva&21.82&\hlc[aqua]{\textbf{51.40}}\\mya\_Mymr&25.23&\hlc[aqua]{\textbf{32.03}}&nld\_Latn&53.61&\hlc[aqua]{\textbf{55.63}}&nya\_Latn&20.34&\hlc[aqua]{\textbf{59.73}}&oci\_Latn&53.78&\hlc[aqua]{\textbf{54.19}}&ory\_Orya&30.57&\hlc[aqua]{\textbf{32.71}}\\pag\_Latn&28.99&\hlc[aqua]{\textbf{41.78}}&scn\_Latn&20.87&\hlc[aqua]{\textbf{24.27}}&slv\_Latn&39.19&\hlc[aqua]{\textbf{51.92}}&sna\_Latn&42.16&\hlc[aqua]{\textbf{47.41}}&snd\_Arab&25.15&\hlc[aqua]{\textbf{36.20}}\\szl\_Latn&33.62&\hlc[aqua]{\textbf{34.22}}&tpi\_Latn&39.00&\hlc[aqua]{\textbf{45.93}}&tsn\_Latn&28.13&\hlc[aqua]{\textbf{56.48}}&tum\_Latn&39.36&\hlc[aqua]{\textbf{60.29}}&vie\_Latn&43.75&\hlc[aqua]{\textbf{52.84}}\\
\noalign{\vskip 1mm}  
\hline
\end{tabular}
\caption{\label{mt1}
Comparison between GPT-3.5-TURBO and EPA. Results in chrF++ for MT on the FLORES-200 dataset. The best results are bolded and highlighted. We report on translating from English into the languages (3-shot on GPT).}
\end{table*}
\begin{table*}[thb!]
\tiny
\centering
    \setlength\tabcolsep{5pt}
\setlength\aboverulesep{0pt}\setlength\belowrulesep{0pt}
\setcellgapes{0pt}\makegapedcells
\begin{tabular}{l|cc|l|cc|l|cc|l|cc|l|cc}
\hline
\noalign{\vskip 1mm}  
\textbf{Language} & \textbf{Copy-9} & \textbf{EPA} & \textbf{Language} & \textbf{Copy-9} & \textbf{EPA} & \textbf{Language} & \textbf{Copy-9} & \textbf{EPA} & \textbf{Language} & \textbf{Copy-9} & \textbf{EPA} & \textbf{Language} & \textbf{Copy-9} & \textbf{EPA}\\
\noalign{\vskip 1mm}  
\hline
\noalign{\vskip 1mm}  
acm\_Arab&15.99&\hlc[aqua]{\textbf{26.28}}&aeb\_Arab&21.12&\hlc[aqua]{\textbf{30.02}}&apc\_Arab&30.38&\hlc[aqua]{\textbf{33.26}}&bjn\_Arab&1.77&\hlc[aqua]{\textbf{2.38}}&dan\_Latn&56.59&\hlc[aqua]{\textbf{64.02}}\\dyu\_Latn&9.07&\hlc[aqua]{\textbf{17.77}}&dyu\_Latn&9.07&\hlc[aqua]{\textbf{17.77}}&dzo\_Tibt&2.44&\hlc[aqua]{\textbf{34.06}}&fij\_Latn&28.31&\hlc[aqua]{\textbf{31.76}}&fin\_Latn&59.49&\hlc[aqua]{\textbf{75.32}}\\hat\_Latn&45.88&\hlc[aqua]{\textbf{56.81}}&hat\_Latn&45.88&\hlc[aqua]{\textbf{56.81}}&isl\_Latn&56.26&\hlc[aqua]{\textbf{59.90}}&kaz\_Cyrl&52.09&\hlc[aqua]{\textbf{58.58}}&kea\_Latn&32.23&\hlc[aqua]{\textbf{34.07}}\\
\noalign{\vskip 1mm}  
\hline
\end{tabular}
\caption{\label{mt2}
Comparison between Copy-9 and EPA. Results in chrF++ for MT on the FLORES-200 dataset. The best results are bolded and highlighted. We report on translating from English into the languages (3-shot on EPA).}
\end{table*}
\subsection{Evaluation Metrics}
\paragraph{Machine Translation} For the evaluation on MT, we report chrF++ scores \citep{popovic-2015-chrf} computed by the sacreBLEU repository.\footnote{https://github.com/mjpost/sacrebleu} We also use COMET \citep{bosselut-etal-2019-comet} with the model version eamt22-cometinho-da.
\paragraph{Dialogue Summarization, Paraphrasing} For the evaluation on dialogue summarization and paraphrasing, we use F1 scores on ROUGE-L \citep{lin-2004-rouge} and BLEU.
\paragraph{Natural Language Inference} We use accuracy to evaluate the performance on the task of NLI.
\subsection{Baselines and Prompt Design}
\paragraph{ChatGPT} We experiment with ChatGPT, a multilingual large language model that has shown strong abilities across various NLU and NLG tasks \citep{2023arXiv230304048W}. At the time of writing, this LLM is widely popular. We use the ChatGPT model versioned GPT-3.5-TURBO. We access ChatGPT via the official API through Python.
\paragraph{Copy-9} As one of the baselines, we repeat the original three demonstrations three times to get 9 demonstrations, which is equal in the number of demonstrations to EPA.
\paragraph{Prompt Design} We use the following prompts for each task we conducted in our experiments:
\begin{itemize}
\setlength\itemsep{0em}
\item \textbf{Machine Translation} \textit{Translate the following text from English into [target-lang]: [source]}
\item \textbf{Dialogue Summarization} \textit{Given the following dialogue: [source] Give the dialogue summarization:}
\item \textbf{Paraphrasing} \textit{Given the English input: [source] Give the paraphrase:}
\item \textbf{Natural Language Inference}  \textit{Given the following two sentences: [source] Whether they are neutral, contradiction, or entailment:} 
\end{itemize}
For the dataset of FLORES-200, we use three pairs of demonstrations for in-context learning on our baselines. For the remaining datasets except for paraphrasing, we use one or three pairs of demonstrations on our baselines. We randomly select the demonstrations from the training sets. We use one demonstration for paraphrasing. All results except for MT are averaged from three runs on three different sets of demonstrations.

\subsection{Easy Prompt Augmentation}
We use ChatGPT to create the paraphrases based on the randomly selected demonstrations. We use the following prompt to create the paraphrases: \textit{Paraphrase the following text: [source]}. For FLORES-200, we create three paraphrases for each English instance and we use the NLLB translator\footnote{https://huggingface.co/spaces/Narrativaai/NLLB-Translator} to acquire the paraphrases in the target language, making it multiple sources and multiple targets. For the remaining tasks, we create one paraphrase for each pair of demonstrations using ChatGPT solely. EPA uses the newly created paraphrases as additional demonstrations for in-context learning.
\section{Results}
\subsection{Main Results}
\paragraph{Machine Translation}
\begin{table}
\tiny
\centering
    \setlength\tabcolsep{20pt}
    \setlength\extrarowheight{0pt}
\begin{tabular}{lcc}
\hline
\noalign{\vskip 1mm}  
\textbf{Model} & \textbf{BLEU} & \textbf{ROUGE-L}\\
\noalign{\vskip 1mm}  
\hline
\noalign{\vskip 1mm} 
\multicolumn{3}{c}{\textit{1-shot}}\\
\hline
\noalign{\vskip 1mm} 
GPT & 9.10 $\pm $ 0.21 & 38.22 $\pm $ 0.25 \\
EPA & \textbf{9.51 $\pm $ 0.16} & \textbf{38.98 $\pm $ 0.14}\\
\noalign{\vskip 1mm}  
\hline
\noalign{\vskip 1mm} 
\multicolumn{3}{c}{\textit{3-shot}}\\
\hline
\noalign{\vskip 1mm} 
GPT & 9.51 $\pm $ 0.18 & 38.77 $\pm $ 0.19\\
EPA & \textbf{9.89 $\pm $ 0.12} & \textbf{39.48 $\pm $ 0.27}\\
\noalign{\vskip 1mm}  
\hline
\end{tabular}
\caption{\label{ds}
Evaluations of EPA and GPT on dialogue summarization. The upper half for 3-shot demonstrations on GPT and EPA applies paraphrasing to the demonstrations. The lower half for 5-shot demonstrations on GPT and EPA applies paraphrasing to the demonstrations. Results averaged from three sets of demonstrations. We found EPA always improve GPT on each set of demonstration.
}
\end{table}

\begin{table}
\tiny
\centering
    \setlength\tabcolsep{20pt}
    \setlength\extrarowheight{0pt}
\begin{tabular}{lcc}
\hline
\noalign{\vskip 1mm}  
\textbf{Model} & \textbf{BLEU} & \textbf{ROUGE-L}\\
\noalign{\vskip 1mm}  
\hline
\noalign{\vskip 1mm} 
GPT & 11.22 $\pm $ 0.09 & 28.90 $\pm $ 0.12\\
EPA & \textbf{11.44 $\pm $ 0.07} & \textbf{29.24 $\pm $ 0.06}\\
\hline
\end{tabular}
\caption{\label{pf}
Evaluations of EPA and GPT on paraphrasing. Results averaged from three sets of demonstrations. We found EPA always improve GPT on each set of demonstration.
}
\end{table}
\begin{table}[htb!]
\tiny
\centering
    \setlength\tabcolsep{20pt}
    \setlength\extrarowheight{0pt}
\begin{tabular}{lcc}
\hline
\noalign{\vskip 1mm}  
\textbf{Model} & \textbf{SNLI-Accuracy} & \textbf{MNLI-Accuracy}\\
\noalign{\vskip 1mm}  
\hline
\noalign{\vskip 1mm} 
\multicolumn{3}{c}{\textit{1-shot}}\\
\hline
\noalign{\vskip 1mm} 
GPT & 54.94 $\pm $ 0.56&53.37 $\pm $ 0.38\\
EPA & \textbf{57.48 $\pm $ 0.43}& \textbf{55.14 $\pm $ 0.29}\\
\noalign{\vskip 1mm}  
\hline
\noalign{\vskip 1mm} 
\multicolumn{3}{c}{\textit{3-shot}}\\
\hline
\noalign{\vskip 1mm} 
GPT & 54.80 $\pm $ 0.44 &50.00 $\pm $ 0.42\\
EPA & \textbf{57.57 $\pm $ 0.48}& \textbf{55.01 $\pm $ 0.52}\\
\noalign{\vskip 1mm}  
\hline
\end{tabular}
\caption{\label{nli}
Evaluations of EPA and GPT on natural language inference. The upper half for 3-shot demonstrations on GPT and EPA applies paraphrasing to the demonstrations. The lower half for 5-shot demonstrations on GPT and EPA applies paraphrasing to the demonstrations. Results averaged from three sets of demonstrations. We found EPA always improve GPT on each set of demonstration.
}
\end{table}
Table \ref{mt1} presents the results in chrF++ on 45 languages for translating from English into these languages. These languages range from high-resource languages to low-resource languages. We compare ChatGPT (GPT-3.5-TURBO) with EPA. The results indicate that EPA brings clear improvements. The gains can be large, by up to about 6x chrF++ points on low-resource languages (1.58 to 9.09 for English to Kashmiri written in Devanagari script (kas\_Deva))  and by up to about 3x chrF++ points on high-resource languages (20.34 to 59.3 for English to Nyanja written in Latin script (nya\_Latn)). The improvement is also consistent for different languages written in different scripts. The results in Table \ref{comet2} in the Appendix also show the same trend. EPA reports an average score of -0.557, which clearly exceeds the score of -0.994 reported by the baseline. This indicates that EPA is an effective approach for improving in-context learning on LLMs.

\paragraph{Dialogue Summarization}
Table \ref{ds} presents the results in BLEU and ROUGE-L on GPT and EPA, where EPA shows good improvement with up to 0.79 improvements in F1 scores on ROUGE-L.
\paragraph{Paraphrasing}
Table \ref{pf} presents the results in BLEU and ROUGE-L on GPT and EPA, where EPA shows good improvement compared to GPT.
\paragraph{Natural Language Inference} Table \ref{nli} reports an evaluation of SNLI and MNLI on the task of NLI. We observe a large performance gain with EPA, by up to 5.01 accuracy on MNLI with 3-shot in-context learning. This indicates the effectiveness of EPA that enhances in-context learning on NLU tasks with LLMs.
\subsection{Analysis - Copy Demonstration}
Table \ref{mt2} reports chrF++ scores on 15 randomly selected languages from NLLB-200 with Copy-9. We observe that copying the demonstrations does not surpass EPA. The performance difference can be large, by up to about 15x difference (2.44 to 34.06 for English to Dzongkha written in Tibetan script (dzo\_Tibt)). We postulate that copying the demonstration can lead to `overfitting' on in-context learning and makes the prompting less generalizable. In contrast, EPA creates paraphrasings in different surface forms and hence improves the generalization of the in-context demonstration learning.
\section{Related Work and Conclusions}
 In-context learning effectively improves downstream task performance. However, most of the scientific research constraints to the situation where those demonstrations are always available \citep{2022arXiv220212837M,2023arXiv230100234D}. This then limits the practical use of in-context learning. Meanwhile, prior works \citep{gao-etal-2020-paraphrase,lu-lam-2023-pcc} have shown that conducting paraphrasing for data augmentation effectively improves the neural language model during the model training stage.  Combining these motivations, we propose a novel method called EPA (\textbf{E}asy \textbf{P}rompt \textbf{A}ugmentation) that effectively reduces user efforts in writing demonstrations while improving the model performance simultaneously. EPA achieves these goals by augmenting demonstrations with multiple sources/targets, where each of them paraphrases each other.  Extensive experiments indicate that EPA effectively improves both NLU and NLG tasks, covering from natural language inference to machine translation in translating tens of languages. The in-depth analysis also indicates that EPA surpasses naively copying the demonstrations.
\section*{Limitations}
This paper presents an analysis of around 40 languages only on MT. However, there are more than thousands of languages around the world. We leave more investigations to our future work. EPA requires an automatic paraphraser to be effective. We do not perform an investigation on the situation where human paraphrasers are available. EPA is also less useful when there exist many demonstrations to be used. Finally, we also conduct experimentation on ChatGPT, which could affect reproducibility in the future.
\section*{Ethical Statement}
We honour and support the ACL Code of Ethics. There is no ethical issue known to us in this work. A well-known and widely used LLM is used in our work, which is subjected to generating offensive context. Yet the above-mentioned issues are widely known to commonly exist for LLMs. Any content generated does not reflect the view of the authors.

\bibliography{ref}

\begin{thebibliography}{23}
\expandafter\ifx\csname natexlab\endcsname\relax\def\natexlab#1{#1}\fi

\bibitem[{Bawden et~al.(2020)Bawden, Zhang, Yankovskaya, T{\"a}ttar, and
  Post}]{bawden-etal-2020-study}
Rachel Bawden, Biao Zhang, Lisa Yankovskaya, Andre T{\"a}ttar, and Matt Post.
  2020.
\newblock \href {https://doi.org/10.18653/v1/2020.findings-emnlp.82} {A study
  in improving {BLEU} reference coverage with diverse automatic paraphrasing}.
\newblock In \emph{Findings of the Association for Computational Linguistics:
  EMNLP 2020}, pages 918--932, Online. Association for Computational
  Linguistics.

\bibitem[{Bosselut et~al.(2019)Bosselut, Rashkin, Sap, Malaviya, Celikyilmaz,
  and Choi}]{bosselut-etal-2019-comet}
Antoine Bosselut, Hannah Rashkin, Maarten Sap, Chaitanya Malaviya, Asli
  Celikyilmaz, and Yejin Choi. 2019.
\newblock \href {https://doi.org/10.18653/v1/P19-1470} {{COMET}: Commonsense
  transformers for automatic knowledge graph construction}.
\newblock In \emph{Proceedings of the 57th Annual Meeting of the Association
  for Computational Linguistics}, pages 4762--4779, Florence, Italy.
  Association for Computational Linguistics.

\bibitem[{Bowman et~al.(2015)Bowman, Angeli, Potts, and
  Manning}]{bowman-etal-2015-large}
Samuel~R. Bowman, Gabor Angeli, Christopher Potts, and Christopher~D. Manning.
  2015.
\newblock \href {https://doi.org/10.18653/v1/D15-1075} {A large annotated
  corpus for learning natural language inference}.
\newblock In \emph{Proceedings of the 2015 Conference on Empirical Methods in
  Natural Language Processing}, pages 632--642, Lisbon, Portugal. Association
  for Computational Linguistics.

\bibitem[{Brown et~al.(2020)Brown, Mann, Ryder, Subbiah, Kaplan, Dhariwal,
  Neelakantan, Shyam, Sastry, Askell, Agarwal, Herbert-Voss, Krueger, Henighan,
  Child, Ramesh, Ziegler, Wu, Winter, Hesse, Chen, Sigler, Litwin, Gray, Chess,
  Clark, Berner, McCandlish, Radford, Sutskever, and
  Amodei}]{NEURIPS2020_1457c0d6}
Tom Brown, Benjamin Mann, Nick Ryder, Melanie Subbiah, Jared~D Kaplan, Prafulla
  Dhariwal, Arvind Neelakantan, Pranav Shyam, Girish Sastry, Amanda Askell,
  Sandhini Agarwal, Ariel Herbert-Voss, Gretchen Krueger, Tom Henighan, Rewon
  Child, Aditya Ramesh, Daniel Ziegler, Jeffrey Wu, Clemens Winter, Chris
  Hesse, Mark Chen, Eric Sigler, Mateusz Litwin, Scott Gray, Benjamin Chess,
  Jack Clark, Christopher Berner, Sam McCandlish, Alec Radford, Ilya Sutskever,
  and Dario Amodei. 2020.
\newblock \href
  {https://proceedings.neurips.cc/paper_files/paper/2020/file/1457c0d6bfcb4967418bfb8ac142f64a-Paper.pdf}
  {Language models are few-shot learners}.
\newblock In \emph{Advances in Neural Information Processing Systems},
  volume~33, pages 1877--1901. Curran Associates, Inc.

\bibitem[{{Dong} et~al.(2022){Dong}, {Li}, {Dai}, {Zheng}, {Wu}, {Chang},
  {Sun}, {Xu}, {Li}, and {Sui}}]{2023arXiv230100234D}
Qingxiu {Dong}, Lei {Li}, Damai {Dai}, Ce~{Zheng}, Zhiyong {Wu}, Baobao
  {Chang}, Xu~{Sun}, Jingjing {Xu}, Lei {Li}, and Zhifang {Sui}. 2022.
\newblock \href {https://doi.org/10.48550/arXiv.2301.00234} {{A Survey on
  In-context Learning}}.
\newblock \emph{arXiv e-prints}, page arXiv:2301.00234.

\bibitem[{Gao et~al.(2020)Gao, Zhang, Ou, and Yu}]{gao-etal-2020-paraphrase}
Silin Gao, Yichi Zhang, Zhijian Ou, and Zhou Yu. 2020.
\newblock \href {https://doi.org/10.18653/v1/2020.acl-main.60} {Paraphrase
  augmented task-oriented dialog generation}.
\newblock In \emph{Proceedings of the 58th Annual Meeting of the Association
  for Computational Linguistics}, pages 639--649, Online. Association for
  Computational Linguistics.

\bibitem[{Gliwa et~al.(2019)Gliwa, Mochol, Biesek, and
  Wawer}]{gliwa-etal-2019-samsum}
Bogdan Gliwa, Iwona Mochol, Maciej Biesek, and Aleksander Wawer. 2019.
\newblock \href {https://doi.org/10.18653/v1/D19-5409} {{SAMS}um corpus: A
  human-annotated dialogue dataset for abstractive summarization}.
\newblock In \emph{Proceedings of the 2nd Workshop on New Frontiers in
  Summarization}, pages 70--79, Hong Kong, China. Association for Computational
  Linguistics.

\bibitem[{{Gong} et~al.(2022){Gong}, {Li}, {Feng}, {Wu}, and
  {Kong}}]{2022arXiv221008933G}
Shansan {Gong}, Mukai {Li}, Jiangtao {Feng}, Zhiyong {Wu}, and Lingpeng {Kong}.
  2022.
\newblock \href {https://doi.org/10.48550/arXiv.2210.08933} {{DiffuSeq:
  Sequence to Sequence Text Generation with Diffusion Models}}.
\newblock \emph{arXiv e-prints}, page arXiv:2210.08933.

\bibitem[{Lin(2004)}]{lin-2004-rouge}
Chin-Yew Lin. 2004.
\newblock \href {https://aclanthology.org/W04-1013} {{ROUGE}: A package for
  automatic evaluation of summaries}.
\newblock In \emph{Text Summarization Branches Out}, pages 74--81, Barcelona,
  Spain. Association for Computational Linguistics.

\bibitem[{Lin et~al.(2022)Lin, Mihaylov, Artetxe, Wang, Chen, Simig, Ott,
  Goyal, Bhosale, Du, Pasunuru, Shleifer, Koura, Chaudhary, O{'}Horo, Wang,
  Zettlemoyer, Kozareva, Diab, Stoyanov, and Li}]{2021arXiv211210668L}
Xi~Victoria Lin, Todor Mihaylov, Mikel Artetxe, Tianlu Wang, Shuohui Chen,
  Daniel Simig, Myle Ott, Naman Goyal, Shruti Bhosale, Jingfei Du, Ramakanth
  Pasunuru, Sam Shleifer, Punit~Singh Koura, Vishrav Chaudhary, Brian O{'}Horo,
  Jeff Wang, Luke Zettlemoyer, Zornitsa Kozareva, Mona Diab, Veselin Stoyanov,
  and Xian Li. 2022.
\newblock \href {https://aclanthology.org/2022.emnlp-main.616} {Few-shot
  learning with multilingual generative language models}.
\newblock In \emph{Proceedings of the 2022 Conference on Empirical Methods in
  Natural Language Processing}, pages 9019--9052, Abu Dhabi, United Arab
  Emirates. Association for Computational Linguistics.

\bibitem[{Lu and Lam(2023)}]{lu-lam-2023-pcc}
Hongyuan Lu and Wai Lam. 2023.
\newblock \href {https://aclanthology.org/2023.eacl-main.5} {{PCC}:
  Paraphrasing with bottom-k sampling and cyclic learning for curriculum data
  augmentation}.
\newblock In \emph{Proceedings of the 17th Conference of the European Chapter
  of the Association for Computational Linguistics}, pages 68--82, Dubrovnik,
  Croatia. Association for Computational Linguistics.

\bibitem[{{Min} et~al.(2022){Min}, {Lyu}, {Holtzman}, {Artetxe}, {Lewis},
  {Hajishirzi}, and {Zettlemoyer}}]{2022arXiv220212837M}
Sewon {Min}, Xinxi {Lyu}, Ari {Holtzman}, Mikel {Artetxe}, Mike {Lewis},
  Hannaneh {Hajishirzi}, and Luke {Zettlemoyer}. 2022.
\newblock \href {https://doi.org/10.48550/arXiv.2202.12837} {{Rethinking the
  Role of Demonstrations: What Makes In-Context Learning Work?}}
\newblock \emph{arXiv e-prints}, page arXiv:2202.12837.

\bibitem[{NLLB-Team(2022)}]{nllb2022}
NLLB-Team. 2022.
\newblock No language left behind: Scaling human-centered machine translation.

\bibitem[{Popovi{\'c}(2015)}]{popovic-2015-chrf}
Maja Popovi{\'c}. 2015.
\newblock \href {https://doi.org/10.18653/v1/W15-3049} {chr{F}: character
  n-gram {F}-score for automatic {MT} evaluation}.
\newblock In \emph{Proceedings of the Tenth Workshop on Statistical Machine
  Translation}, pages 392--395, Lisbon, Portugal. Association for Computational
  Linguistics.

\bibitem[{Rubin et~al.(2022)Rubin, Herzig, and
  Berant}]{rubin-etal-2022-learning}
Ohad Rubin, Jonathan Herzig, and Jonathan Berant. 2022.
\newblock \href {https://doi.org/10.18653/v1/2022.naacl-main.191} {Learning to
  retrieve prompts for in-context learning}.
\newblock In \emph{Proceedings of the 2022 Conference of the North American
  Chapter of the Association for Computational Linguistics: Human Language
  Technologies}, pages 2655--2671, Seattle, United States. Association for
  Computational Linguistics.

\bibitem[{{Tang} et~al.(2023){Tang}, {Lu}, {Jiang}, {Huang}, {Zhang}, {Zhao},
  and {Wei}}]{2023arXiv230515067T}
Tianyi {Tang}, Hongyuan {Lu}, Yuchen~Eleanor {Jiang}, Haoyang {Huang}, Dongdong
  {Zhang}, Wayne~Xin {Zhao}, and Furu {Wei}. 2023.
\newblock \href {https://doi.org/10.48550/arXiv.2305.15067} {{Not All Metrics
  Are Guilty: Improving NLG Evaluation with LLM Paraphrasing}}.
\newblock \emph{arXiv e-prints}, page arXiv:2305.15067.

\bibitem[{Thompson and Post(2020)}]{thompson-post-2020-automatic}
Brian Thompson and Matt Post. 2020.
\newblock \href {https://doi.org/10.18653/v1/2020.emnlp-main.8} {Automatic
  machine translation evaluation in many languages via zero-shot paraphrasing}.
\newblock In \emph{Proceedings of the 2020 Conference on Empirical Methods in
  Natural Language Processing (EMNLP)}, pages 90--121, Online. Association for
  Computational Linguistics.

\bibitem[{{Wang} et~al.(2023){Wang}, {Liang}, {Meng}, {Shi}, {Li}, {Xu}, {Qu},
  and {Zhou}}]{2023arXiv230304048W}
Jiaan {Wang}, Yunlong {Liang}, Fandong {Meng}, Haoxiang {Shi}, Zhixu {Li},
  Jinan {Xu}, Jianfeng {Qu}, and Jie {Zhou}. 2023.
\newblock \href {https://doi.org/10.48550/arXiv.2303.04048} {{Is ChatGPT a Good
  NLG Evaluator? A Preliminary Study}}.
\newblock \emph{arXiv e-prints}, page arXiv:2303.04048.

\bibitem[{Wei et~al.(2022)Wei, Wang, Schuurmans, Bosma, brian ichter, Xia, Chi,
  Le, and Zhou}]{2022arXiv220111903W}
Jason Wei, Xuezhi Wang, Dale Schuurmans, Maarten Bosma, brian ichter, Fei Xia,
  Ed~H. Chi, Quoc~V Le, and Denny Zhou. 2022.
\newblock \href {https://openreview.net/forum?id=_VjQlMeSB_J} {Chain of thought
  prompting elicits reasoning in large language models}.
\newblock In \emph{Advances in Neural Information Processing Systems}.

\bibitem[{Wei and Zou(2019)}]{wei-zou-2019-eda}
Jason Wei and Kai Zou. 2019.
\newblock \href {https://doi.org/10.18653/v1/D19-1670} {{EDA}: Easy data
  augmentation techniques for boosting performance on text classification
  tasks}.
\newblock In \emph{Proceedings of the 2019 Conference on Empirical Methods in
  Natural Language Processing and the 9th International Joint Conference on
  Natural Language Processing (EMNLP-IJCNLP)}, pages 6382--6388, Hong Kong,
  China. Association for Computational Linguistics.

\bibitem[{Williams et~al.(2018)Williams, Nangia, and Bowman}]{N18-1101}
Adina Williams, Nikita Nangia, and Samuel Bowman. 2018.
\newblock \href {http://aclweb.org/anthology/N18-1101} {A broad-coverage
  challenge corpus for sentence understanding through inference}.
\newblock In \emph{Proceedings of the 2018 Conference of the North American
  Chapter of the Association for Computational Linguistics: Human Language
  Technologies, Volume 1 (Long Papers)}, pages 1112--1122. Association for
  Computational Linguistics.

\bibitem[{{Zhang} et~al.(2022){Zhang}, {Roller}, {Goyal}, {Artetxe}, {Chen},
  {Chen}, {Dewan}, {Diab}, {Li}, {Lin}, {Mihaylov}, {Ott}, {Shleifer},
  {Shuster}, {Simig}, {Singh Koura}, {Sridhar}, {Wang}, and
  {Zettlemoyer}}]{2022arXiv220501068Z}
Susan {Zhang}, Stephen {Roller}, Naman {Goyal}, Mikel {Artetxe}, Moya {Chen},
  Shuohui {Chen}, Christopher {Dewan}, Mona {Diab}, Xian {Li}, Xi~Victoria
  {Lin}, Todor {Mihaylov}, Myle {Ott}, Sam {Shleifer}, Kurt {Shuster}, Daniel
  {Simig}, Punit {Singh Koura}, Anjali {Sridhar}, Tianlu {Wang}, and Luke
  {Zettlemoyer}. 2022.
\newblock \href {https://doi.org/10.48550/arXiv.2205.01068} {{OPT: Open
  Pre-trained Transformer Language Models}}.
\newblock \emph{arXiv e-prints}, page arXiv:2205.01068.

\bibitem[{Zhang et~al.(2022)Zhang, Feng, and Tan}]{zhang-etal-2022-active}
Yiming Zhang, Shi Feng, and Chenhao Tan. 2022.
\newblock \href {https://aclanthology.org/2022.emnlp-main.622} {Active example
  selection for in-context learning}.
\newblock In \emph{Proceedings of the 2022 Conference on Empirical Methods in
  Natural Language Processing}, pages 9134--9148, Abu Dhabi, United Arab
  Emirates. Association for Computational Linguistics.

\end{thebibliography}
\bibliographystyle{acl_natbib}
\appendix
\section{COMET Scores}
\begin{table*}[thb!]
\tiny
\centering
    \setlength\tabcolsep{6pt}
\setlength\aboverulesep{0pt}\setlength\belowrulesep{0pt}
\setcellgapes{0pt}\makegapedcells
\begin{tabular}{l|cc|l|cc|l|cc|l|cc|l|cc}
\hline
\noalign{\vskip 1mm}  
\textbf{Language} & \textbf{GPT} & \textbf{EPA} & \textbf{Language} & \textbf{GPT} & \textbf{EPA} & \textbf{Language} & \textbf{GPT} & \textbf{EPA} & \textbf{Language} & \textbf{GPT} & \textbf{EPA} & \textbf{Language} & \textbf{GPT} & \textbf{EPA}\\
\noalign{\vskip 1mm}  
\hline
\noalign{\vskip 1mm}  
acm\_Arab&-0.96&\hlc[aqua]{\textbf{-0.22}}&afr\_Latn&-1.13&\hlc[aqua]{\textbf{-0.52}}&als\_Latn&-0.96&\hlc[aqua]{\textbf{-0.71}}&arz\_Arab&-1.14&\hlc[aqua]{\textbf{0.05}}&asm\_Beng&-1.07&\hlc[aqua]{\textbf{-0.23}}\\azb\_Arab&-1.04&\hlc[aqua]{\textbf{-0.74}}&bak\_Cyrl&-0.99&\hlc[aqua]{\textbf{-0.08}}&bjn\_Arab&-1.13&\hlc[aqua]{\textbf{0.68}}&crh\_Latn&-0.76&\hlc[aqua]{\textbf{-0.51}}&dan\_Latn&-0.98&\hlc[aqua]{\textbf{-0.49}}\\eus\_Latn&-0.83&\hlc[aqua]{\textbf{-0.31}}&ewe\_Latn&-1.13&\hlc[aqua]{\textbf{-0.34}}&fin\_Latn&-0.84&\hlc[aqua]{\textbf{-0.39}}&fuv\_Latn&-0.71&\hlc[aqua]{\textbf{-0.43}}&gaz\_Latn&-0.97&\hlc[aqua]{\textbf{-0.35}}\\guj\_Gujr&-0.86&\hlc[aqua]{\textbf{0.42}}&jpn\_Jpan&-0.52&\hlc[aqua]{\textbf{-0.12}}&kas\_Deva&-1.11&\hlc[aqua]{\textbf{-0.13}}&khk\_Cyrl&-0.95&\hlc[aqua]{\textbf{0.43}}&khm\_Khmr&-1.11&\hlc[aqua]{\textbf{-0.84}}\\kik\_Latn&-0.99&\hlc[aqua]{\textbf{-0.32}}&kir\_Cyrl&-1.36&\hlc[aqua]{\textbf{-0.42}}&kmb\_Latn&-0.63&\hlc[aqua]{\textbf{-0.29}}&kor\_Hang&-0.88&\hlc[aqua]{\textbf{-0.39}}&lij\_Latn&-0.70&\hlc[aqua]{\textbf{-0.33}}\\lim\_Latn&-1.02&\hlc[aqua]{\textbf{-0.17}}&lmo\_Latn&-1.14&\hlc[aqua]{\textbf{-0.27}}&ltg\_Latn&-1.11&\hlc[aqua]{\textbf{-0.29}}&lug\_Latn&-1.37&\hlc[aqua]{\textbf{-0.59}}&mai\_Deva&-1.01&\hlc[aqua]{\textbf{-0.43}}\\mya\_Mymr&-0.99&\hlc[aqua]{\textbf{-0.55}}&nld\_Latn&-1.10&\hlc[aqua]{\textbf{0.42}}&nya\_Latn&-1.05&\hlc[aqua]{\textbf{-0.13}}&oci\_Latn&-1.10&\hlc[aqua]{\textbf{0.25}}&ory\_Orya&-0.83&\hlc[aqua]{\textbf{-0.50}}\\pag\_Latn&-1.32&\hlc[aqua]{\textbf{0.20}}&scn\_Latn&-1.14&\hlc[aqua]{\textbf{-0.31}}&slv\_Latn&-1.02&\hlc[aqua]{\textbf{-0.48}}&sna\_Latn&-0.91&\hlc[aqua]{\textbf{-0.32}}&snd\_Arab&-0.95&\hlc[aqua]{\textbf{-0.17}}\\szl\_Latn&\hlc[aqua]{\textbf{-0.84}}&-0.91&tpi\_Latn&-1.11&\hlc[aqua]{\textbf{-0.05}}&tsn\_Latn&-1.04&\hlc[aqua]{\textbf{-0.28}}&tum\_Latn&-1.07&\hlc[aqua]{\textbf{-0.25}}&vie\_Latn&-0.99&\hlc[aqua]{\textbf{0.02}}\\
\noalign{\vskip 1mm}  
\hline
\end{tabular}
\caption{\label{comet2}
Comparison between GPT-3.5-TURBO and EPA. Results in COMET for MT on the FLORES-200 dataset. The best results are bolded and highlighted. We report on translating from English into the languages (3-shot on GPT).}
\end{table*}

\end{document}